# Housing Market Prediction Problem using Different Machine Learning Algorithms: A Case Study

Shashi Bhushan Jha[1], Radu F. Babiceanu[1, *], Vijay Pandey[2], Rajesh Kumar Jha[3]

[1]Department of Electrical Engineering and Computer Science, Embry-Riddle Aeronautical University, Daytona Beach, FL 32114, USA
[2]Department of Computer Science Engineering, IIT Kharagpur, India
[3]Department of Electronics and Communication Engineering, BNMIT, India

E-mail: jhas1@my.erau.edu, babicear@erau.edu, vijayiitkgp13@gmail.com, rajeshjnv23@gmail.com

*Corresponding author (phone: +1-386-226-7535; email: babicear@erau.edu)

**Abstract**

Developing an accurate prediction model for housing prices is always needed for socio-economic development and well-being of citizens. In this paper, a diverse set of machine learning algorithms such as XGBoost, CatBoost, Random Forest, Lasso, Voting Regressor, and others, are being employed to predict the housing prices using public available datasets. The housing datasets of 62,723 records from January 2015 to November 2019 is obtained from the Florida's Volusia County Property Appraiser website. The records are publicly available and include the real estate/economic database, maps, and other associated information. The database is usually updated weekly according to the State of Florida regulations. Then, the housing price prediction models using machine learning techniques are developed and their regression model performances are compared. Finally, an improved housing price prediction model for assisting the housing market is proposed. Particularly, a house seller/buyer or a real estate broker can get insight in making better-informed decisions considering the housing price prediction. The empirical results illustrate that based on prediction model performance, Coefficient of Determination ($R^2$), Mean Square Error (*MSE*), Mean Absolute Error (*MAE*), and computational time, the XGBoost algorithm performs superior than the other models to predict the housing price.

## 1) Introduction

Starting with 2005, the increasing interest rates in the U.S. housing market have slowed down the market considerably. Particularly, the investment bank Lehman Brothers Holdings was affected significantly, and forced into bankruptcy in 2008. This resulted in a sharp decline in the housing prices and, combined with the subprime mortgage crisis, increased the slowing down of the economy and weakened the asset values, which ultimately led to the depreciation of the global housing market and caused a global crisis (Park & Kwon Bae, 2015).

Consequently, economists turned their attention to predicting these types of threats that could jeopardize the economic stability.

After the 2008 global crisis, the housing market fell for several years, especially in the large cities, until the end of 2011. Beginning with 2012, the housing market followed an upward trend with decreasing inventories, increasing demand, and naturally increasing prices. This again made economists and market analysers turn their attention to more precise prediction models to shield the economy from predictable threats that could result in economic downturns (Park & Kwon Bae, 2015).

Machine learning has been used in prediction for some time now with increasingly better results that were put in practice and changed the economic landscape. Practically every economic domain now benefits from machine learning prediction models, and the current models are becoming more accurate given the computational power available for processing immense sets of data. In this research, the housing price problem is analysed using several machine learning techniques such as XGBoost, CatBoost, Lasso, Voting Regressor, Random Forest, Decision Tree, Linear Regression, and Support Vector Regressor. First, we propose a target binning technique based on machine learning to solve the housing market problem. An extensive computational experiment has been performed on the housing problem considering a distinct set of features, to develop a prediction model with high $R^2$ score and low *MSE* and *MAE* values. The solutions obtained by using the XGBoost algorithm using target binning for the housing price problem outperforms in comparison with model score and computation time those obtained by using the other models.

The rest of the paper is structured as follows. Section 2 covers the housing price prediction literature. Then, Section 3 presents the case study for the housing price problem, and the proposed housing problem modeling framework, including data analysis and processing. Section 4 presents the machine learning techniques that have been applied for the housing price problem, followed by the empirical results of the models and a comparison of the results in Section 5. Finally, the conclusions of the study and the future work are outlined in Section 6.

## 2) Literature Review

The reviewed literature was divided in four separate parts. First, the studies that emphasize then housing price evaluation using machine learning techniques are reviewed. The second part includes the studies focusing on hedonic-based regression, and other stochastic approaches for the price prediction problem. The third part of the literature review concentrates on the studies related to price prediction model using specifically machine learning algorithms. Finally, the last part of this section uncovers the identified research gap and the contributions of this study.

### 2.1. Studies on the housing problem and machine learning techniques

Park and Bae (2015) addressed the house price prediction problem considering the housing data available for the Virginia's Fairfax County. To solve the problem, the authors have employed machine learning techniques such as Naive Bayesian, AdaBoost, and RIPPER to develop a house price classification model. Their reported results demonstrated that the RIPPER algorithm performs better than all other models. In their study, the authors suggested as future scope to consider the appraised value of a property, property tax, and also to increase their limited dataset size. Our current research considered those recommendations and all the three factors have been included in our research for the Volusia county dataset to build a more robust model and enhance the existing literature. In another study, Gu et. al. (2011) studied the housing price problem with the aim of forecasting a house price model. A hybrid of genetic algorithms and support vector machines method was proposed to solve the model. The model dealt with a housing dataset that was collected in China, during the 1993-2002 period. In the end, the results showed that the used G-SVM approach performed better than the grey model. Plakandaras et. al. (2015) also addressed the U.S. real estate house price index problem. In their research, a novel hybrid forecasting method was proposed combining the ensemble empirical mode decomposition (EEMD) with Support Vector Regression (SVR). The obtained solutions of their proposed model are compared with Random Walk (RW), Bayesian Vector Autoregressive, and Bayesian Autoregressive models.

*2.2. Studies on the price prediction problem using hedonic and other stochastic approaches*

He and Xia (2020) studied the housing price problem stressing on heterogeneous traders and a healthy urban housing market. Their paper covered the speculative investment effects on house price and economic disturbance and proposed a dynamic stochastic general equilibrium model to solve the problem. The solutions of the problem describe that the quick increase and moderate decrease of house price negatively affect a healthy house market. The study is directed towards sustainable economic development and affirms that property tax needs to be implemented properly considering market-oriented reforms. Ceritoglu et. al. (2019) investigated the regional house price problem in Turkey to detect unsustainable market enthusiasm and prevent collapsing of the market. The research analyses the real estate hedonic house price and rent price in 26 geographical areas of the country. To solve the problem, a right tailed unit root testing method was considered. The empirical results show that the multiple periods were characterized by false enthusiasm episodes in house prices. As a result, drastically increasing price behavior led to market implosion in several regions beginning in 2018. Law (2017) explored a street based local area (SLA) problem and measured the effects of house prices employing hedonic price procedure. This empirical problem was a case study of the metropolitan area of London to estimate hedonic price models and to identify local house submarkets. The author considered a multi-level hedonic price method for testing the conjecture and concluded that SLA has a notable effect on house prices. The study also showed that there are considerable local area effects on housing price, which resulted in an overall recommendation that SLA be preferred in comparison with larger region-based models for housing price problems.

*2.3. Studies on price prediction problem using machine learning approaches*

Nam and Seong (2019) studied stock market prediction problems by analysing media housing market information considering unstructured data and utilizing the asymmetric relationships of firms. To solve this problem, the authors proposed a machine learning model to predict stock price movement with the help of financial news incorporating casualty. For this problem, a Korean market dataset was considered for the experimental results. The solutions obtained by this study show better performance compared to traditional machine learning techniques. In another research, the prediction of the daily bitcoin exchange rate was been considered and the behavior of financial markets was studied (Mallqui & Fernandes, 2019). Authors proposed a method including the machine learning features to solve the bitcoin exchange rate prediction problem. The results of this problem were validated with the solutions of state-of-the-art papers of this wide research area. Díaz et. al. (2019) considered the prediction problem of Spanish day-ahead electricity price. To solve the problem, a regression tree-based approach has been proposed. Moreover, in this problem the dataset, particularly, the model variables are obtained from publicly accessible energy consumption datasets. The results of the model illustrate reasonable accuracy for price formation prediction and provides argumentation for the use of non-linear analysis to predict the price using independent variables.

*2.4. Research gap and contribution*

In the existing literature, a limited amount of work has been focused on the housing price prediction model, particularly, to solve the problem using machine learning approaches. A few identified papers were reported above. In addition, most of the past research considered the housing market problem as classification problems to develop a classification model instead of a regression model. Therefore, the objective of the study is to predict the housing price valuation using machine learning techniques and considering competitive regression models. An improved ML based algorithm is proposed, which includes the predicted target price binning variable as features in the model and improves the model accuracy significantly. More precisely, the model accuracy is increased by 10 percent compared to other contemporary machine learning techniques. Moreover, to the best of our knowledge, the analysed datasets available from the Volusia County Property Appraiser have not been used in any previous research of housing price prediction problems.

**3) Case Study and Modeling Framework**

In this section, a general overview of the case study and a real-world case of house pricing problem is presented. In addition, this section includes the information coming from the property sales datasets of the Volusia County Property Appraiser and the proposed modeling framework to analyse the datasets.

*3.1. Problem Description and Research Framework*

The primary aim of this case study is to predict housing price for the given features to maximize the prediction accuracy by employing the proposed methodology. This housing problem can be considered both as a regression or a classification housing problem. Since the classification problem was previously reported in the literature, this research considers several regression models with target variable binning which are applied on the housing market data to predict the property price. Fig. 1 captures the research framework for the housing price prediction problem. It includes five major blocks, namely data collection, data preparation, feature processing, model training, and model evaluation. These blocks of the diagram are explained in detail in the next subsections.

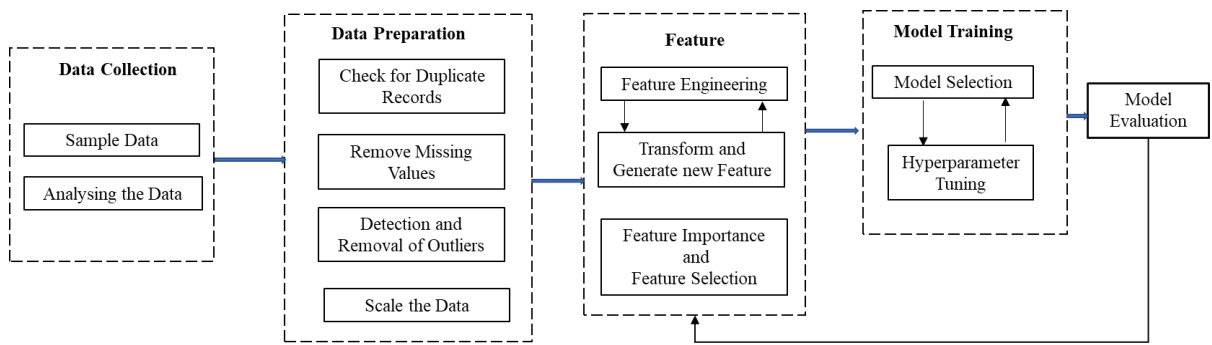

Fig. 1. Research framework for the housing price problem

*3.2. Data Description and Collection*

For this research, the real estate datasets have been gathered from the Volusia County Property Appraiser website[1] which is publicly available. The datasets include an updated property sale database, maps, and other associated information. The database is weekly updated according to Florida Statutes and Florida Department of Revenue Substantive Regulations. Particularly, the property data and sales record are collected including the listed and closed prices of the property. In this study, the last five years of housing data within the January 1, 2015, to November 13, 2019 period, with a specific qualification of the property is considered. Initially, the total 62,723 records with 19 variables were extracted from the database. For this research, the dataset is maintained using the open-source PostgreSQL relation database management system.

*3.3. Data Preparation and Cleaning*

The second block of the research framework of Fig. 1 focuses on data preparation. In the considered dataset, few records were accommodated with duplicate entries or null values. These duplicate entries were eliminated, and the records of null values were removed from the dataset in the preparation stage. Moreover, adding missing values to the dataset is not considered in this study, because the number of entries is high enough and thus it is better to remove the missing value entries altogether. Some features were excluded from the dataset because of their low correlations with the target variable, or lack of contribution to the increase the house price prediction accuracy, such as the zip code. The correlation matrix of the dataset generated by Pearson Correlation is shown in Fig. 2. After the data preparation process, the remaining total observations left in the dataset is 50,809 records. Table 1 includes the description of features and target variable with the mean and standard deviation of each variable. It can be observed that most of the features are real data types which are significantly required for the regression model. But there are also few features of object type or date type still part of the dataset such as sale date and zip code. For the visualization of the collected and prepared data, a Violin plot of target variable is presented in Fig. 3. The plot shows the price distribution and helps in obtaining a good insight of the actual data used in the regression models. Also for visualization purposes, the *t*-SNE, *t*-distributed Stochastic Neighbor Embedding (Maaten & Hinton, 2008) is employed in Fig. 4, and shows that data is complex and not linearly separable.

---

[1] http://vcpa.vcgov.org/

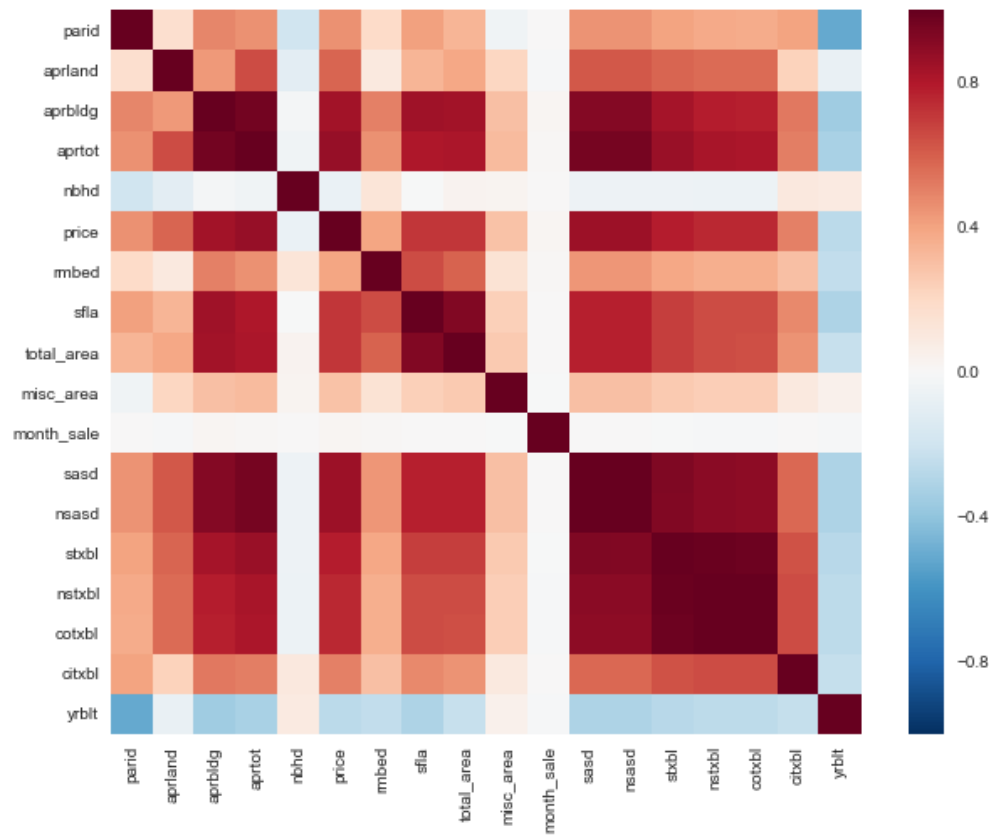

Fig. 2. Correlation matrix for the given dataset generated by Pearson Correlation

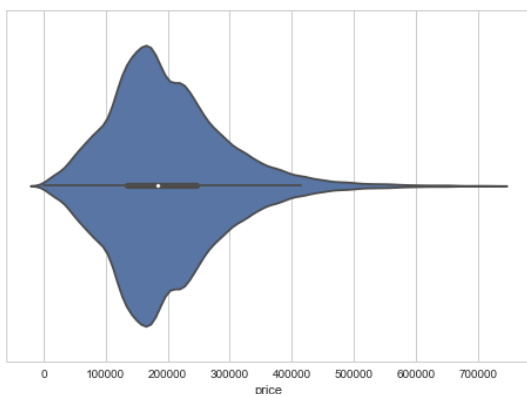
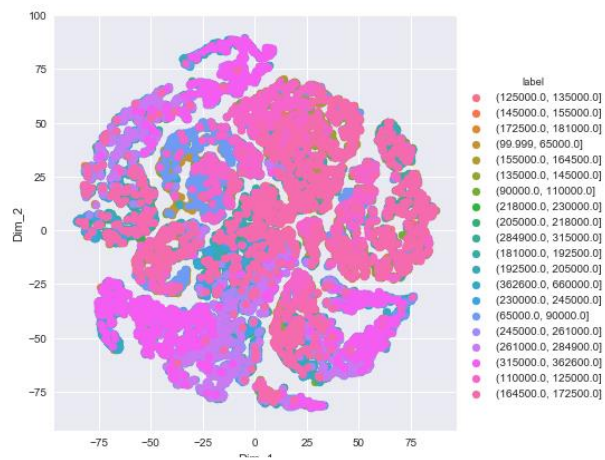

Fig. 3. Price distribution on Violin Plot    Fig. 4. All feature dimensions presented in 2D

*3.3.1. Removing Outliers*

In data analytics, as part of the data cleaning, outlier detection and removal help in making the training model more stable. Outliers may make the model unstable and result in an increase in the variance. In Fig. 5, the Cook's influence diagram (Cook, 1979) is employed, and it can be noticed that a few data points are highly influential and have the capability to change the estimator value drastically. Thus, it is advisable to remove these data points first. In this study, two outlier detection and removal methods, $z$-squared and Tukey Fences, are applied to find which one performs better in detecting and removing the outliers of the dataset.

Table 1. Description of the variables

| Variable Name | Description | Mean | Std. Deviation |
|---|---|---|---|
| *Parid* | Parcel ID (property ID) | 4386556 | 1780110 |
| *Aprland* | Total Land Just Value | 31512.52238 | 22474.45819 |
| *Aprbldg* | Total Building(s) Just Value | 153352.0601 | 65863.09319 |
| *Aprtot* | Just Value at time of Sale or Total Just Value | 184864.5824 | 78255.48143 |
| *Nbhd* | Neighborhood Code | 3553.249727 | 1617.738503 |
| *Rmbed* | Number of bedrooms | 2.940746 | 0.747859 |
| *Sfla* | Square Footage of Living Area | 1670.539912 | 590.574028 |
| *total_area* | Total Building Square Footage | 2409.546287 | 817.763307 |
| *yrblt* | Year Built | 1988 | 21 |
| *misc_area* | Miscellaneous area that includes the gym, swimming pool, parking | 141.641915 | 236.598402 |
| *ZIP21* | zip code of area | 32394.25951 | 289.84512 |
| *sale_date* | sale date of parcel | - | - |
| *sasd* | School Assessed Value | 172068.0029 | 75437.46026 |
| *nsasd* | Non-School Assessed Value | 171783.2722 | 75619.59938 |
| *stxbl* | School Taxable | 154036.8721 | 78289.80939 |
| *nstxbl* | Non-School Taxable Value | 139935.5139 | 79855.06776 |
| *cotxbl* | County Taxable Value | 138761.7173 | 80899.04927 |
| *citxbl* | Sale Price of House City Taxable Value | 109971.8981 | 88619.63572 |
| *Price* | Sale Price of House | 197912.7255 | 94021.27873 |

First, the $z$-score is used for the outlier removal process. Data points having $z$-score value between -3 and +3 are considered. After employing the $z$-score based technique, 10% data points are detected as outliers. The second approach using the Tukey Fences based method has also been employed to determine which of the two techniques is more appropriate for the dataset. The equations of Tukey Fences (Rousseeuw, Ruts, & Tukey, 1999) are as follows:

$$Q1 = CDF^{-1}(a) \qquad (1)$$

$$Q3 = CDF^{-1}(b) \qquad (2)$$

$$IQR = Q2 - Q1 \qquad (3)$$

where, $CDF^{-1}$ is the quantile function (cumulative distribution function), and $IQR$ is the interquartile range. Outliers are detected as the data points observed that fall below $Q1 - 1.5 \times IQR$ or above $Q3 + 1.5 \times IQR$. Usually the values of $a$ and $b$ are selected as 0.25 and 0.75, respectively, however in this paper, the values of $a$ and $b$ are selected as 0.10 and 0.90, respectively. After applying the above two methods, it has been found that the $IQR$-based method, Tukey Fences, is more appropriate for the given dataset.

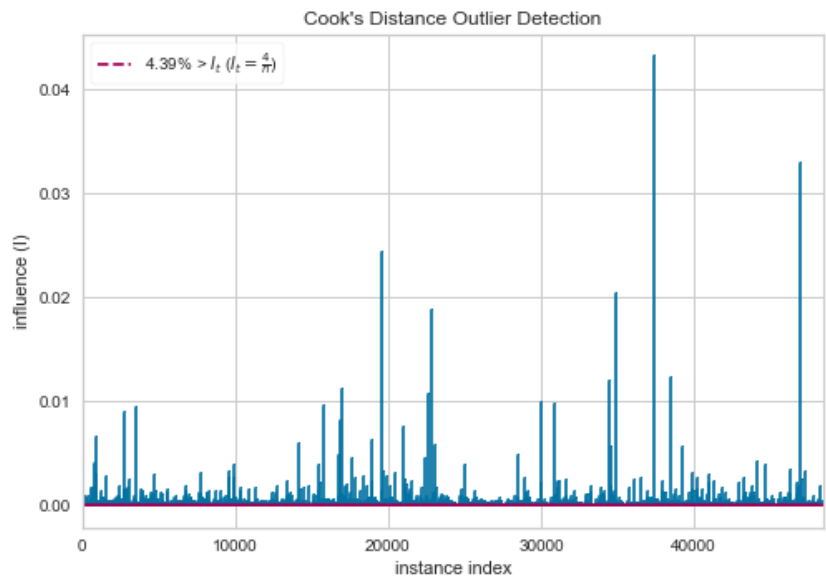

Fig. 5. Cook's Distance outlier detection for the given property dataset

### 3.2.2. Removing Duplications

For regression modeling, if some data points are replicated by being present more than once in the dataset, they are more strongly represented in the underlying data, so the regression algorithm treats them as having more importance. It can be thought of each occurrence of a data point as pulling the regression line towards it with the same force. If there are two data points at a given point in the regression model plane, they will pull the line towards them twice as hard. Therefore, it is indicated to remove the duplicate values from data itself. Duplication removal should be done carefully though, as these duplicate data points may cause what is called data leakage, when splitting the entire data into training and validation sets. If a data point in the training set has a duplicate value in the validation set, then the model will give biased prediction toward these duplicate data points, which is not desirable since will bias the entire prediction model. The duplicate entries and null entries from the housing sale transactions dataset were removed. Also, since the number of entries in the dataset is large, there is no need to replace the missing values. Those entries were removed altogether.

*3.3. Feature Engineering*

The main purpose of feature engineering is to find the most influential or partially important features of the dataset and detect the less valuable features to be removed from the dataset. The process results in a highly efficient and less complex model. Following this process requires domain expert knowledge in identifying a set of key features of the dataset and performing feature analysis, which is described in the next subsection.

*3.3.1. Feature Analysis*

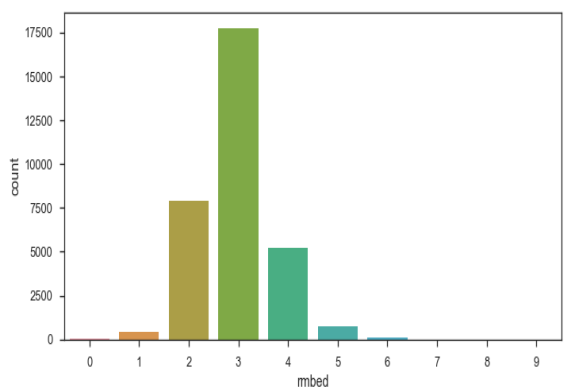 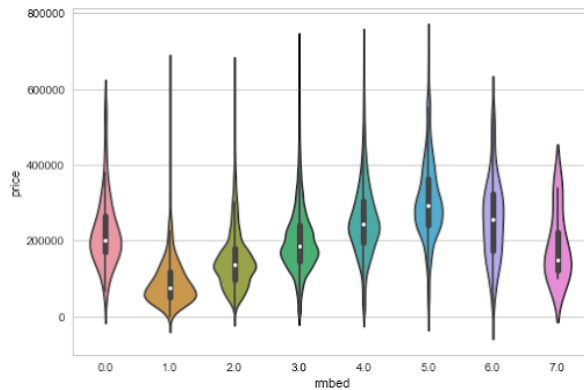

Fig. 6. Number of bedrooms per house   Fig. 7. Violin Plot between bedrooms and property sale price

Fig. 6 illustrates that most buyers considered in the dataset purchase a house with two or three bedrooms. Therefore, most of the residents in the purchased houses occupy it with nuclear families having a small number of people. The subsequent figure (Fig. 7) depicts the direct relationship between the increase in the number of bedrooms and the increase in the house sale price extracted from the dataset. However, at the highest end, for houses with 6 or 7 bedrooms the direct relationship does not stand. This behavior could be explained by the location of the properties, which likely are found in a rural environment, considerably outside of city locations. Then, Fig. 8 depicts the seasonality of housing purchase, showing a peak of transactions in the May to August months. This peak sale could be explained by the summer vacation. On the contrary, the November to February months are characterized by lower number of housing sale transactions.

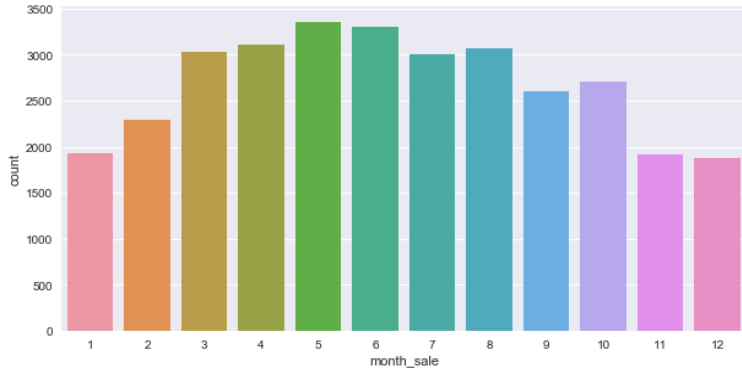

Fig. 8. Property sale transactions in each month

*3.3.2. Feature Transformation and Target Binning*

A couple of the features of the dataset are selected for transformation: the sale date and the year built. For this process, the feature encoding is used, and the sale date and year-built features are converted into *ys*1 and *ys*2, and *yb*1 and *yb*2, respectively. The feature transformation process is followed by the target binning process on the sale price. The steps of the target binning techniques are as follows:

**Step-1:** At the beginning, a *price_range* column of 100 bin is generated in the dataset that contains the values between 0 to 100. Further, after using the level encoder, the feature from categorical to numerical feature is encoded and stored as *price_bin*.

**Step-2:** In step 2, the price is dropped from the dataset, with *price_bin* taken as target variable and used for training the model. After performing the model with *price_bin* as target variable, the *R*-square score of the model obtained is 0.91 by employing the same model (*e.g.*, XGBoost), and predicting the *price_bin* value for each observation of the dataset (*i.e.*, 48,469 records).

**Step-3:** In step 3, the predicted *price_bin* values are fed to the dataset as input, as independent variables. Then, all the outliers of the newly generated *price_bin* feature are removed using *IQR*. Using this process, a total of 660 records are deleted from the dataset. Finally, the actual price value is taken as target variable and the model is trained using the XGBoost algorithm. As a result, the $R^2$ value of the model increased significantly from 0.92 to 0.97.

*3.3.3. Feature Importance and Selection*

Identifying feature importance is performed on the given dataset having 19 variables by employing distinct machine learning techniques. Generally, this process needs to be performed with care. When less data is available, it becomes difficult to generalize the model well because in most scenarios few features do not represent data very well. As mentioned in the previous section domain expert knowledge is needed for identifying a set of key features. The domain expert selects away the non-important features and only the important or partially important features are kept for further processing. For the considered housing dataset, tree-based feature importance methods have been used to gain insight of the available data and identify the important and partially important features for each machine learning technique considered. Fig. 9, 10, and 11 present this process for the Random Forest Regressor and XGBoost Regressor techniques.

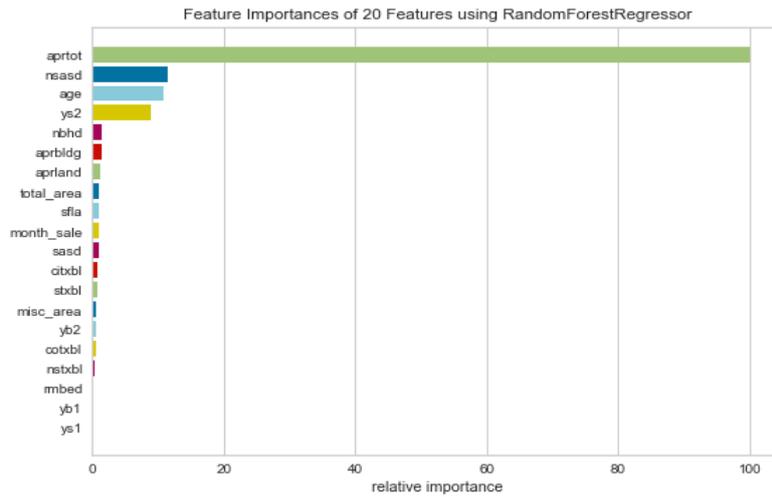

Fig. 9. Feature importance of 20 features using Random Forest Regressor

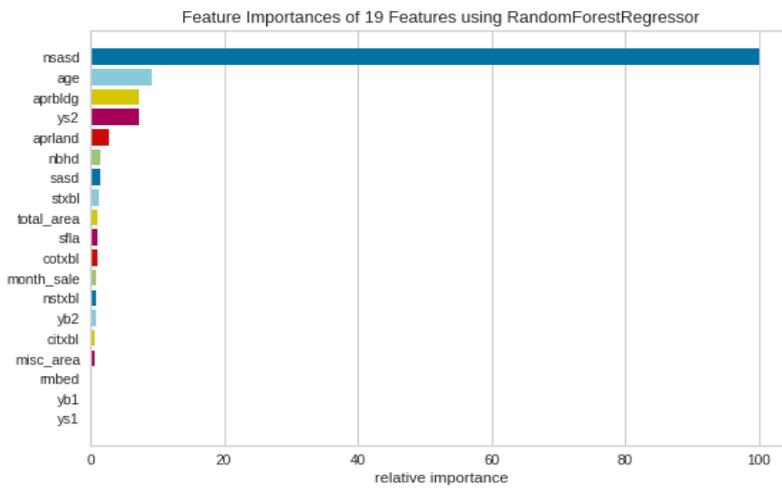

Fig. 10. Feature importance after excluding *aprtot* variable from the dataset using Random Forest Regressor

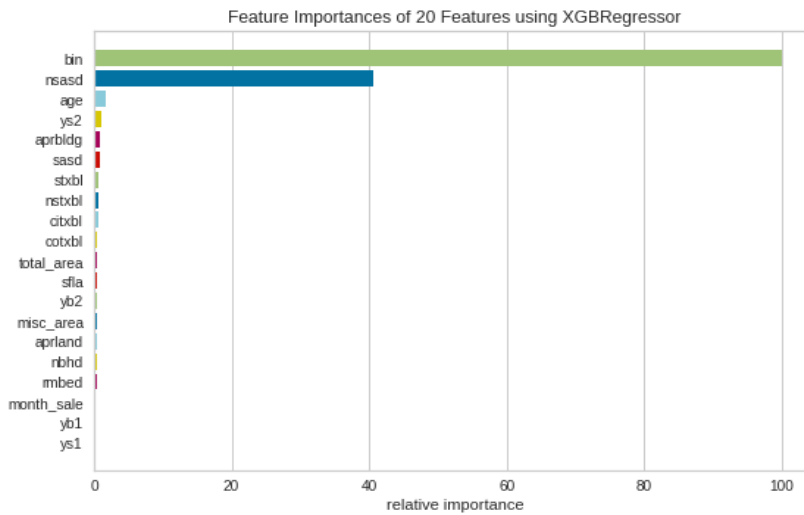

Fig. 11. Feature importance of 20 features using XGBoost Regressor

**4) Machine Learning Methodologies Employed**

The *t*-SNE chart of Fig. 4, presented in a previous section, shows that the considered dataset is complex and not linearly separable. Few algorithms can be used on complex data, such as Support Vector Networks (Cortes & Vapnik, 1995) and tree-based methods. For tree-based methods, available algorithms include XGBoost (Chen & Guestrin, 2016), Random Forest (Breiman, 2001), and Gradient Boosting (Ridgeway, 2007), which perform well on complex data with better $R^2$ value after proper hyper parameter tuning. Besides the above algorithms, this study also employs the CatBoost algorithm (Dorogush, Ershov, & Gulin, 2018; Prokhorenkova et. al. 2018), which is very helpful to handle categorical features. The category variables for the processed dataset are month sale, *nbhd*, and zip code. Various ensemble techniques (Breiman, 1996; Louppe & Geurts, 2012) have been used to remove the bias of individual estimators, and to improve the data variance. Voting regressor (An & Meng, 2010) has also been used, as part of ensemble technique. For the selected algorithms, the number of features was kept as low as possible, to reduce the overfitting problem and to increase the interpreting capability. Adding non-important features creates unwanted noise in the data, and as mentioned previously it was avoided.

*4.1. CatBoost Method*

CatBoost (Dorogush et al., 2018; Prokhorenkova et al., 2018) is a recently published machine learning technique that is based on gradient boosted decision trees (Ridgeway, 2007). When the model consumes the dataset and starts training the model, a set of decision trees (Breiman et. al., 1984) is built consecutively. Each successive tree is built with reduced loss compared to the previous trees. Moreover, in this model, the number of trees is controlled by the defined parameters. To prevent overfitting, this model uses the overfitting detector that resides in the algorithm. As the detector is triggered, the decision trees are stopped being built.

*4.2. Random Forest Method*

Random forest (Breiman, 2001) is an ensemble of decision trees where each tree is built from a sample drawn from the training set. To give more randomness in building a random forest, some random subset of given features or all features are considered for best split, while splitting operations on each node (Ho, 1998). Size of the random subset is passed by the user as a hyper parameter. The individual decision tree suffers from high variance problems that lead to overfitting of the tree estimator. Random forest overcomes the problem of high variance in individual tree by providing above-mentioned two types of randomness. Random subset samples make different errors, and thus estimators generalize well by taking the uniform average of each predictor that helps in cancelling out the errors. Generally, random forests suffer from the increased bias problem, but variance is the key point to take care over bias.

*4.3. Lasso Method*

The Lasso (Tibshirani, 1996) is a linear model that is mostly used for feature elimination by making coefficients of some non-important features to zero. Mathematically, the objective function of Lasso is defined by a linear model with an added regularization term. The Lasso objective function is given in equation (4):

$$\min_{w} \frac{1}{2n_{samples}} ||Xw - y||_2^2 + \alpha ||w||_1 \qquad (4)$$

The objective is to minimize the least square penalty with $\alpha ||w||_1$ added, where $\alpha$ is a constant and $||w||_1$ is the $l_1$-norm of the coefficient vector.

*4.4. XGBoost Method*

XGBoost (Chen & Guestrin, 2016) stands for "Extreme Gradient Boosting" and is a technique based on the concept of gradient boosting trees (Friedman, 2001). The main difference from other gradient boosting based techniques is the objective function, which consists of two parts: training loss and regularization term, as presented in the equation (5).

$$\mathcal{L}(\emptyset) = \sum_i \ell(\hat{y}_i, y_i) + \sum_k \Omega(f_k) \qquad (5)$$

$$\text{where, } \Omega(f) = \gamma T + \frac{1}{2}\lambda ||w||^2$$

The training loss measures how predictive the model is with respect to the training data. The regularization term controls the complexity of the model that helps to improve the model generalization. A common choice of training loss is the mean squared error. In case of XGBoost, the Taylor expansion of the loss function up to the second order is used to expand the polynomials loss function.

*4.5. Voting Regressor Method*

Voting regressor (An & Meng, 2010) works on the concept of combining different machine learning techniques and returning the uniform average predicted values. A voting regressor is a technique that uses base regressors and fits each of them on the whole dataset. Such a regressor is useful for a set of equally well performing estimators in order to balance out their individual weaknesses. Ensemble methods perform best when the predictors are as independent from one another as possible. Generally, the solution is to use a different algorithm to train each regressor to make each predictor more independent from each other. This increases the chance they will make different types of errors, improving the ensemble's performance.

*4.6. Model Construction and Evaluation*

For regression problems, the model performance metric is based on the coefficient of determination, $R^2$, and the Mean Square Error (*MSE*) and Mean Absolute Error (*MAE*) values. Usually, $R^2$ value is placed between 0 and 1. The lower values indicates that the model exhibits none to minimal variability of the response data around its mean. The higher values of the coefficient denote the model has large variability of the response data around its mean. In addition, *MSE* is considered because outliers have been already removed thus the model assumption is that no outlier in the data exists. For instance, if any outlier lies in the data then it will be penalized higher in *MSE*. The smaller value of *MSE* designates the lower average errors of the prediction, and the better performance of the model. The *k*-fold cross validation technique is used to compare the extrapolation ability of a model. Model will be tested on 10 disjoint validation sets and their average is calculated to determine the final score.

For the feature selection method, this study adopts 5-fold cross validation on the training data to decide the relative optimal set of predictor variables. The process starts with all variables selected for model. Then, the feature importance method is used to obtain the important variables until the evaluation criteria of internal cross validation reaches the maximum. Finally, the selected subset of variables is used in the outer evaluation. During the model construction, certain machine learning algorithms have hyper-parameters that need to be tuned, such as the tree depth, regularization parameter for XGBoost, number of trees and sample size for each split in Random Forest, regularization alpha value in Lasso regression. For the parameters, the grid search method is further applied in the innermost 5-fold cross validation, and the parameters with optimal $R^2$ value are selected to train the model.

The specific technique employed to build these machine learning models and tune the meta-parameters considers the Scikit-learn package from the Python platform (Pedregosa et al., 2011). While training and testing the model, the Yellowbrick visualization tool, a Scikit-learn package (Bengfort & Bilbro, 2019) tool is used. To better visualize the performance of the models, two visualizing approaches are considered: (1) Residuals Plot is used to visualize the difference between residuals on the vertical axis and the dependent variable on the horizontal axis; (2) The Prediction Error Plot is used for visualizing the prediction errors as a scatterplot of the predicted and actual values. The line of best fit is visualized and compared to the 45° line.

**5) Machine Learning Empirical Results**

This section summarizes the empirical results of the target binning method with machine learning techniques applied on the selected housing problem to determine the price prediction model. After data cleaning and feature engineering on the dataset, the presented machine learning techniques are employed on the data. In particular, the model is trained using the Random Forest algorithm without incorporating the target binning and the $R^2$ value of the model achieves 0.89. Using the procedure presented in the previous sections, then the model is trained using the same algorithm but this time with target binning. The resultant performance of the model improves this way to 0.97. The training procedure for the house pricing model is continued with several other algorithms. The results of running these models are presented in the following subsections and their performance comparison summarized in Table 2.

*5.1. Support Vector Regression and Linear Regression Techniques Performance*

The Linear Regression (LR) and Support Vector Regression (SVR) algorithms were employed to train the model without including the target binning method. Their resultant scores obtained were 0.80 and 0.78, respectively. The performance of the model does not achieve satisfactory level and the mean square error of the models is very high. This implies the data have many outliers. Thus, removal of outliers is performed on the price binning using the *IQR* and then the same model incorporating the predicted target binning as feature is applied. The score of the model improves significantly to 0.97 for LR and 0.87 for SVR, as seen in Table 2.

### 5.2. Random Forest Technique Performance

Random Forest (RDF) is a machine learning algorithm that usually produces good results even without hyper-parameter tuning. It preserves the substantial score of a large amount of data and it also deals with the missing values efficiently. The chances of overfitting are low, obtaining a good score of $R^2$ after training. The results of the model are depicted in Fig. 12 below, which clearly indicate that there is no overfitting, with the $R^2$ of the RDF model achieving 0.97 that is 10% more than SVR model. Fig. 13 compares the model fit line with the best fit line (*i.e.*, the 45º line) and shows that the prediction error is clearly lower. Moreover, the residuals plot, which calculated the error of the prediction, depicts that the difference between the target and predicted value is significantly lower that can be seen in Fig. 12.

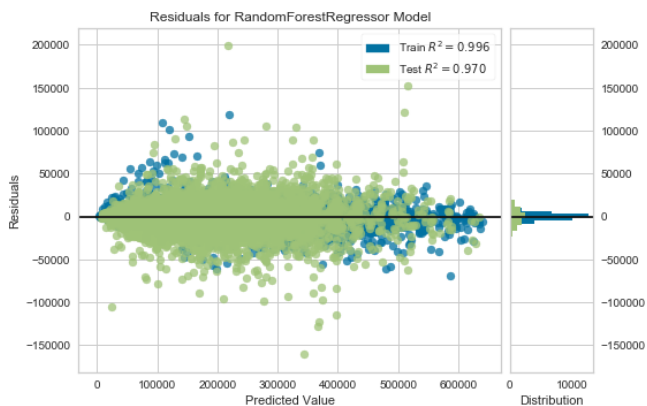
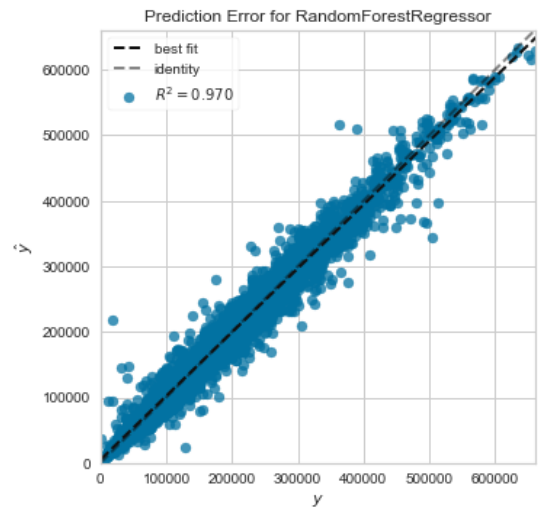

Fig. 12. Residuals for Random Forest model         Fig. 13. Prediction error for Random Forest model

### 5.3. Lasso Regression Technique Performance

For regularization of linear models, the Alpha Selection Visualizer establishes how unique values of alpha impact the model selection. A high value of alpha increases the regularization parameter, which impacts on the model. If alpha value is zero, there is no regularization, so it can be concluded that the alpha factor enhances the effect of regularization. For error minimization, an optimal alpha value needs to be determined. Fig. 14 depicts the plot for the selection of alpha considering the error. For this study, an alpha value of 0.393 is chosen. Same as for the previous model, the plot of residuals and prediction error for the Lasso model are generated and shown in Fig. 15 and 16, respectively. It can be seen that the training and testing score of the Lasso model are 0.969 and 0.971, respectively. The significant $R^2$ score of the model designates the robustness of the model.

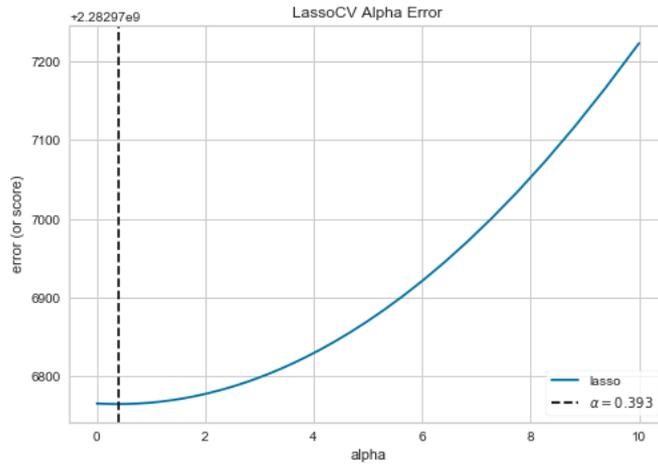

Fig. 14. Lasso alpha error

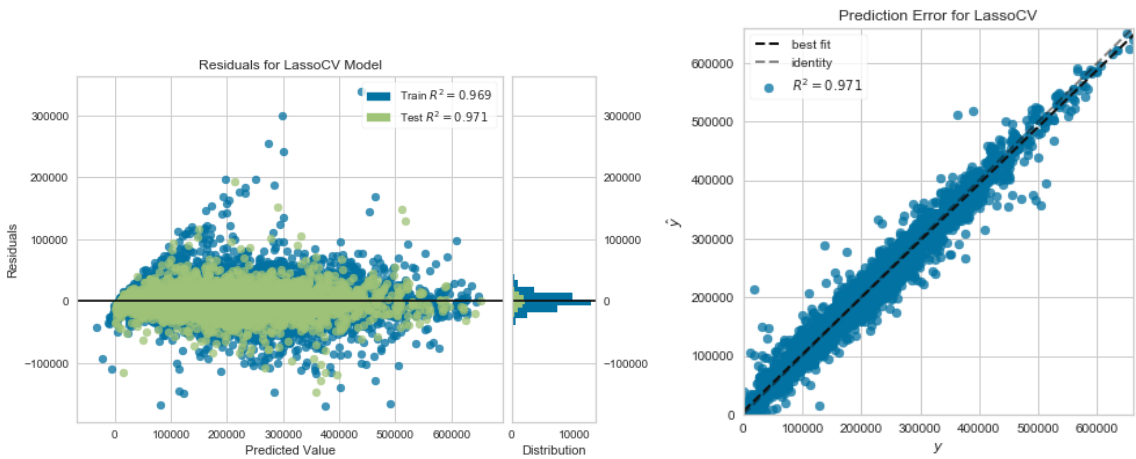

Fig. 15. Residuals for Lasso model　　　　　Fig. 16. Prediction error for Lasso model

### 5.4. XGBoost Regressor Technique Performance

In practice, the XGBoost method is used for boosted tree algorithms. In this study, XGBoost method, using target binning, is employed on the tabular data of the housing problem with the aim to determine the better $R^2$ score and the minimal error rate. With respect to error rate, typically XGBoost has lower performance compared to the RDF algorithm. For the approximation of loss function, the Taylor series is applied up to the second order derivative to improve the model generalization and regularization, The $R^2$ score of this model yields 0.996 and 0.968 for training and testing, respectively. This study prefers the XGBoost over the RDF algorithm even there is a slightly small difference between the $R^2$ scores of the two. The $R^2$ score of XGBoost is 0.968, while the $R^2$ score for RDF is 0.970. However, XGBoost generates a lower *MSE* value of 8.17 compared to an *MSE* of 8.20 generated by the RDF. In addition, the XGBoost allows various options to tune the hyper parameters. To analyze the performance of the prediction model, the prediction error and the residuals for XGBoost regressor are illustrated in Fig. 17 and 18, respectively. Further, Fig. 19 shows that the performance learning curve also converges irrespective of training and testing results.

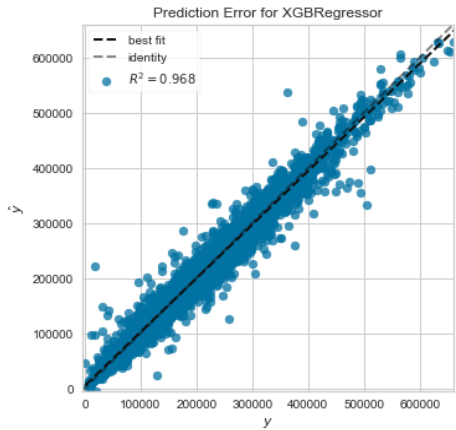
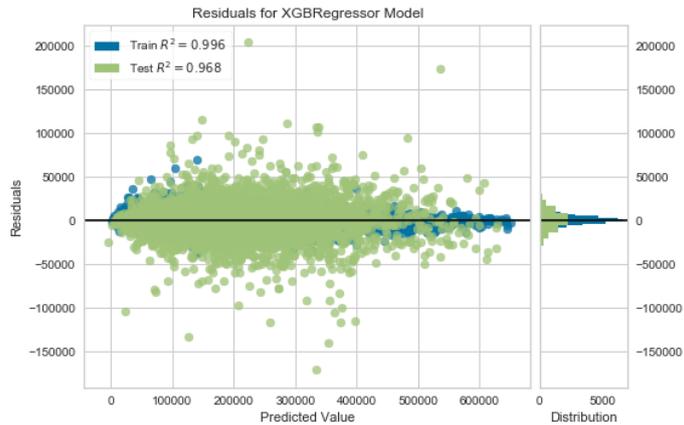

Fig. 17. Prediction error for XGBoost model

Fig. 18. Residuals for XGBoost model

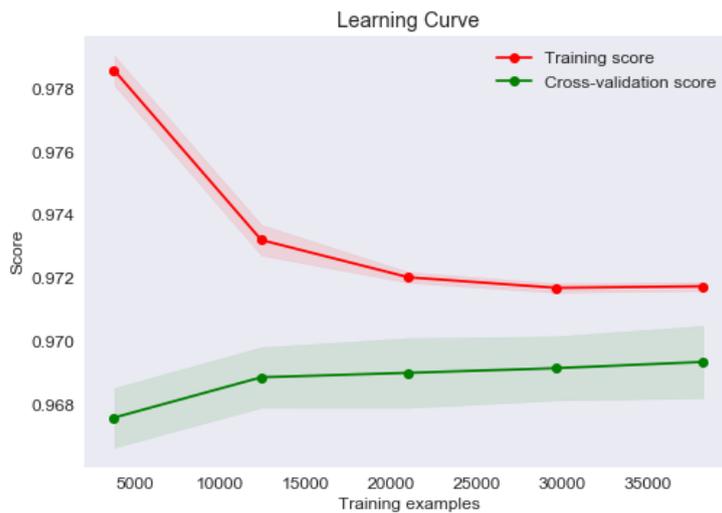

Fig. 19. Performance learning curve of XGBoost model

## 5.5. Voting Regressor Technique Performance

For the Voting Regressor model, three competitive approaches of machine learning algorithms are selected. These approaches have been already applied and received the outputs on the housing problem individually in the above sections, namely, Random Forest, Lasso, and XGBoost. As a general approach, the Voting Regressor model finds the best result by averaging all generated outputs of selected models. After applying this regression model considering target binning on housing price problem, the $R^2$ score values obtained are 0.996 and 0.970, for training and testing models, respectively. The prediction error and residuals plot of Voting Regressor are shown in Fig. 20 and 21, respectively.

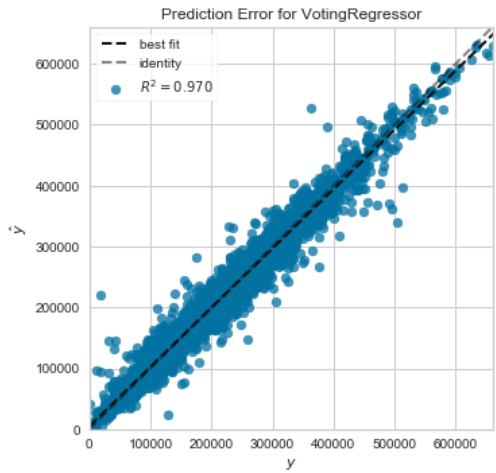 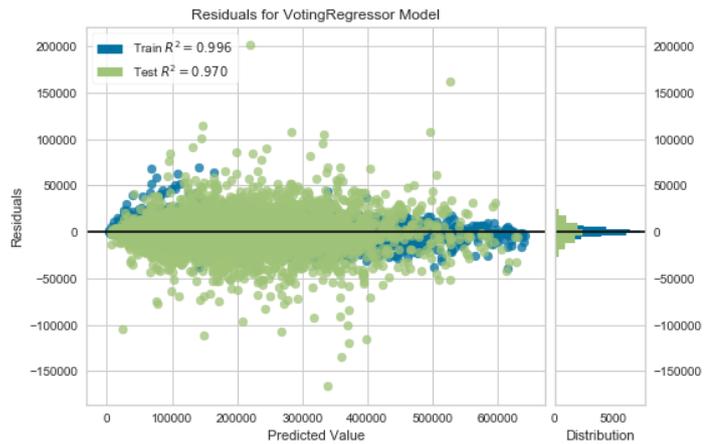

| Fig. 20. Prediction error for Voting Regressor model | Fig. 21. Residuals for Voting Regressor model |

## 5.6. Technique Performance Comparison

Table 2 summarizes the results obtained by applying the above machine learning techniques, before and after price binning, for the housing price prediction problem.

Table 2: Performance comparison before and after price binning on each model

| Sr. No. | Models | Before Target Binning | | | After Target Binning | | |
|---|---|---|---|---|---|---|---|
| | | $R^2$ | MSE | MAE | $R^2$ | MSE | MAE |
| 1 | VCPA Model | 0.74 | 9.36 | 4.51 | - | - | - |
| 2 | Linear Regression | 0.8 | 9.98 | 4.33 | 0.97 | 9.87 | 4.53 |
| 3 | SVR | 0.78 | 9.83 | 4.36 | 0.87 | 9.29 | 4.47 |
| 4 | Decision Tree | 0.84 | 9.1 | 4.36 | 0.93 | 8.57 | 4.13 |
| 5 | Random Forest | 0.91 | 8.79 | 4.2 | 0.97 | 8.2 | 4.07 |
| 6 | XGBoost | 0.92 | 8.59 | 4.1 | 0.97 | 8.17 | 4.03 |
| 7 | Lasso | 0.81 | 9.18 | 4.33 | 0.97 | 8.19 | 4.17 |
| 8 | Voting Regressor | 0.9 | 8.87 | 4.36 | 0.97 | 8.37 | 4.04 |
| 9 | CatBoost | 0.92 | 8.81 | 4.22 | 0.97 | 8.09 | 4.07 |

The appraised price suggested by the Volusia County Property Appraiser (VCPA) and actual sale price of the property are considered to determine the score and robustness of the existing model. This performance of the VCPA model considered as the benchmark for the machine learning study. The VCPA model $R^2$ score value is significantly lower compared to the results of this machine learning study. Also, the error rate is in most of the cases higher than the results of this machine learning study, both before and after target binning.

For the studied housing problem, two approaches are presented by integrating machine learning algorithms with target binning and without target binning. In the first type of experiments without target binning, the results of Table 2 show that the $R^2$ score value of Random Forest, Voting Regressor, XGBoost, and CatBoost models are better than for the other models Linear Regression, SVR, Decision Tree, and Lasso, and all the solutions are more desirable than the existing VCPA model. Then, when using target binning, Table 2 shows that the $R^2$ score value of Linear Regression and SVR models improves drastically by 0.17 and 0.9, respectively. All models are more desirable again than the existing VCPA model.

Comparing the machine learning models with one another, the performance of XGBoost, CatBoost, and Random Forest, when considering target binning, are superior and almost the same for all three metrics ($R^2$ *MSE*, and *MAE*). The models can be differentiated by their computational time, with Random Forest being clearly the slowest, and the XGBoost being the fastest. Thus, this study chooses target binning XGBoost and CatBoost over all other models for the housing price prediction problem. For an enhanced visual comparison of model performances, a total of 50 random test samples are selected from the test dataset and the predicted prices using the trained model are depicted for all models, including the benchmark model. Fig. 22 visualizes these 50 random tests along with the actual house sale price to show the performance of each model.

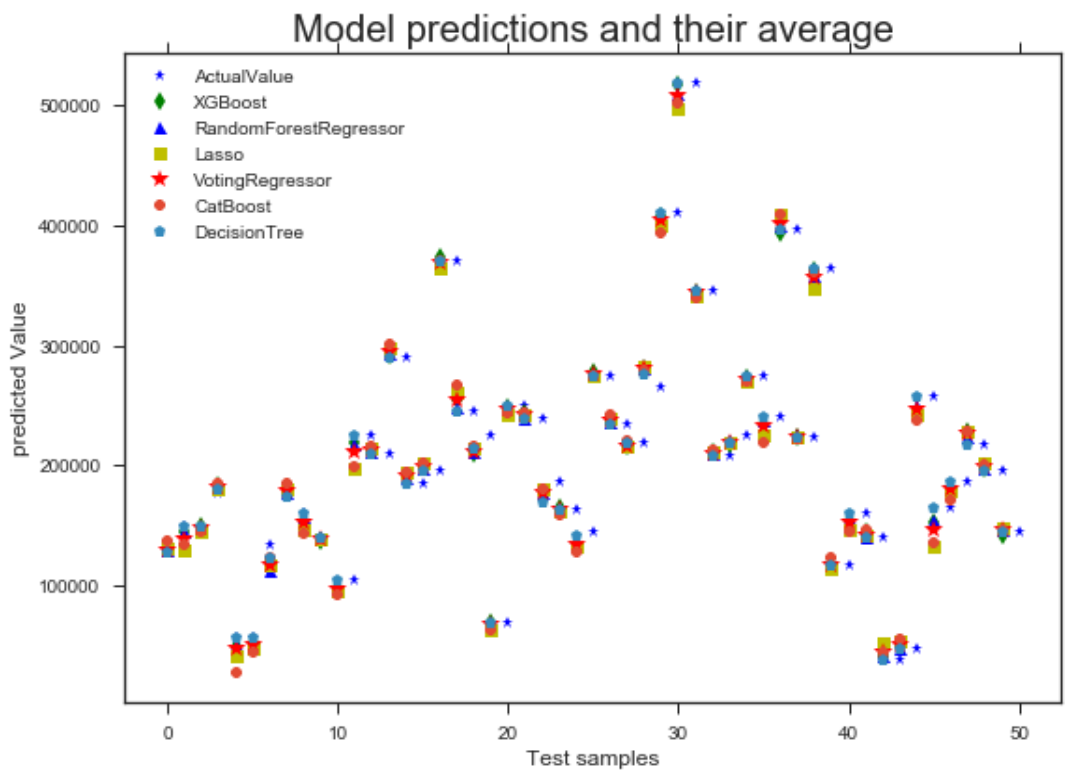

Fig. 22. Model prediction and their average for each model and actual property value

## 6) Conclusions and Future Work

This study employs machine learning techniques, with and without target binning, to develop a price prediction model for housing problems. It uses a rather large publicly available dataset of real estate transactions for a 5-year period. The regression model performances of the models are compared with one another and with the benchmark model. The empirical results show that the XGBoost algorithm with target binning provides superior performance for all metrics under study, the coefficient of determination $R^2$ score, the mean errors, and the computational time.

The developed model may facilitate the prediction of future housing prices and the establishment of policies for the real estate market. Particularly, the sellers and buyers of properties can benefit from this study and make better-informed decisions regarding the property evaluation. In addition, property agents can focus on the seasonality effects, especially during the summer season, when most of the people buy their properties, and on the clear preference for two- or three-bedroom properties. The financial organizations and mortgage lenders may also find the study beneficial and identify more accurate real estate property value, risk analysis, and lending decisions.

The study can be enlarged in a subsequent research by increasing the dataset size so potentially uncovered details and features of the dataset and of this study can be addressed. An increased dataset would potentially be good enough for employing deep neural networks, which can assure that more in-depth analysis on the housing price prediction can be performed. Then, the enlarged housing price prediction problem can be tackled as a classification problem.